\documentclass[11pt,a4paper]{article}
\usepackage[hyperref]{conll2018}
\usepackage{times}
\usepackage{latexsym}

\usepackage{url}

\usepackage{adjustbox}
\usepackage{listings}
\usepackage{caption,setspace}
\usepackage{diagbox} % package for diagonal box in table 
\usepackage{booktabs} % package for toprule, midrule, bottomrule
\usepackage{todonotes}
\usepackage[inline]{enumitem}
\usepackage{amssymb}

\usepackage{todonotes} %%% temp, for revising

\usepackage{color}
\newcommand{\ie}{i.e., }
\newcommand{\figref}[1]{Fig.~\ref{#1}}    % within sentence
\newcommand{\Figref}[1]{Figure~\ref{#1}}  % start of sentence
\newcommand{\tabref}[1]{Table~\ref{#1}}
\newcommand{\Tabref}[1]{Table~\ref{#1}}
\newcommand{\secref}[1]{Section~\ref{#1}}

\usepackage[percent]{overpic}
\usepackage{amsmath}

\usepackage{multirow}

\aclfinalcopy % Uncomment this line for the final submission

\title{Predefined Sparseness in Recurrent Sequence Models}

\author{Thomas Demeester, 
        {\bf Johannes Deleu,} 
        {\bf Fr\'ederic Godin,} 
        {\bf Chris Develder} \\
  Ghent University - imec \\
  Ghent, Belgium\\
  {\tt firstname.lastname@ugent.be} \\}

\date{}

\begin{document}
\maketitle
\begin{abstract}
Inducing sparseness while training neural networks has been shown to yield models with a lower memory footprint but similar effectiveness to dense models. However, sparseness is typically induced starting from a dense model, and thus this advantage does not hold during training. 
We propose techniques to enforce sparseness upfront in recurrent sequence models for NLP applications, to also benefit training.
First, in language modeling, we show how to increase hidden state sizes in recurrent layers without increasing the number of parameters, leading to more expressive models.
Second, for sequence labeling, we show that word embeddings with predefined sparseness lead to similar performance as dense embeddings, at a fraction of the number of trainable parameters.
\end{abstract}

\section{Introduction} 

Many supervised learning problems today are solved with deep neural networks exploiting large-scale labeled data. The computational and memory demands associated with the large amount of parameters of deep models can be alleviated by using \emph{sparse} models.
Applying sparseness can be seen as a form of regularization, as it leads to a reduced amount of model parameters\footnote{The sparseness focused on in this work, occurs on the level of trainable parameters, i.e., we do not consider data sparsity.}, for given layer widths or representation sizes. 
Current successful approaches gradually induce sparseness during training, starting from densely initialized networks, as detailed in \secref{sec:relatedwork}. However, we propose that models can also be built with \emph{predefined sparseness}, \ie such models are already sparse by design and do not require sparseness inducing training schemes.

The main benefit of such an approach is memory efficiency, even at the start of training. 
Especially in the area of natural language processing, in line with the hypothesis by \citet{Yang_2017} that natural language is ``high-rank'', it may be useful to train larger sparse representations, even when facing memory restrictions. 
For example, in order to train word representations for a large vocabulary using limited computational resources, predefined sparseness would allow training larger embeddings more effectively compared to strategies inducing sparseness from dense models.

The contributions of this paper are 
\begin{enumerate*}[label=(\roman*)]
\item a predefined sparseness model for recurrent neural networks,
\item as well as for word embeddings, and
\item proof-of-concept experiments on part-of-speech tagging and language modeling, including an analysis of the memorization capacity of dense vs.\ sparse networks.
\end{enumerate*}
An overview of related work is given in the next \secref{sec:relatedwork}. We subsequently present  predefined sparseness in recurrent layers (\secref{sec:sparse_rnns}), as well as embedding layers (\secref{sec:sparse_emb}), each illustrated by experimental results. This is followed by an empirical investigation of the memorization capacity of language models with predefined sparseness (\secref{sec:L2R}). \secref{sec:conclusion} summarizes the results, and points out potential areas of follow-up research.

The code for running the presented experiments is publically available.\footnote{https://github.com/tdmeeste/SparseSeqModels}

\section{Related Work}\label{sec:relatedwork}

A substantial body of work has explored the benefits of using sparse neural networks.
In deep convolutional networks, common approaches include sparseness regularization, e.g., using decomposition \cite{liu_sparse_2015} or variational dropout \cite{molchanov_variational_2017}), pruning of connections \cite{han_deep_2015, han_learning_2015, guo_dynamic_2016} and low rank approximations \cite{jaderberg_speeding_2014, tai_convolutional_2015}. Regularization and pruning often lead to mostly random connectivity, and therefore to irregular memory accesses, with little practical effect in terms of hardware speedup. Low rank approximations are structured and thus do achieve speedups, with as notable examples the works of \citet{wen_learning_2016} and \citet{lebedev_fast_2016}.

Whereas above-cited papers specifically explored convolutional networks, our work focuses on recurrent neural networks (RNNs). Similar ideas have been applied there, e.g., see \citet{lu_learning_2016} for a systematic study of various new compact architectures for RNNs, including low-rank models, parameter sharing mechanisms and structured matrices. Also pruning approaches have been shown to be effective for RNNs, e.g., by \citet{narang_exploring_2017}. 
Notably, in the area of audio synthesis, \citet{kalchbrenner_2018} showed that 
large sparse networks perform better than small dense networks. Their sparse models were obtained by pruning, and importantly, 
a significant speedup was achieved through an efficient implementation.

For the domain of natural language processing (NLP), recent work by \citet{wang_deep_2016} provides an overview of sparse learning approaches, and in particular noted that ``application of sparse coding in language processing is far from extensive, when compared to speech processing''.
Our current work attempts to further fill that gap. In contrast to aforementioned approaches (that either rely on inducing sparseness starting from a denser model, or rather indirectly try to impose sparseness by enforcing constraints), we explore ways to predefine sparseness. 
%include? 
%https://arxiv.org/pdf/1803.04831.pdf
%https://arxiv.org/pdf/1802.03676.pdf

In the future, we aim to design models where predefined sparseness will allow using very large representation sizes at a limited computational cost. This could be interesting for training models on very large datasets \cite{Chelba2013, Shazeer2017}, or for more complex applications such as joint or multi-task prediction scenarios \cite{Miwa2016, bekoulis2018, hashimoto2017}.

\section{Predefined Sparseness in RNNs}\label{sec:sparse_rnns}
Our first objective is designing a recurrent network cell with fewer trainable parameters than a standard cell, with given input dimension $i$ and hidden state size $h$. 
In \secref{subsec:sparse_rnn}, we describe one way to do this, while still allowing the use of fast RNN libraries in practice. This is illustrated for the task of language modeling in \secref{subsec:LM}.

\subsection{Sparse RNN Composed of Dense RNNs}\label{subsec:sparse_rnn}
The weight matrices in RNN cells can be divided into input-to-hidden matrices $\mathbf{W}_{hi}\in \mathbb{R}^{h\times i}$ and 
hidden-to-hidden matrices $\mathbf{W}_{hh}\in \mathbb{R}^{h\times h}$ (assuming here the output dimension corresponds to the hidden state size $h$), adopting the terminology used in \cite{Goodfellow_2016}. 
A \emph{sparse} RNN cell can be obtained by 
introducing sparseness in 
$\mathbf{W}_{hh}$ and $\mathbf{W}_{hi}$. 
Note that our experiments make use of the Long Short-Term Memory (LSTM) cell \cite{hochreiter_97}, 
but our discussion should hold for any type of recurrent network cell. For example, an LSTM
contains 4 matrices $\mathbf{W}_{hh}$ and $\mathbf{W}_{hi}$, whereas the Gated Recurrent Unit (GRU) \cite{Chung2014_GRU} only has 3.

We first propose 
to organize the hidden dimensions in several disjoint groups, i.e, $N$ segments with lengths $s_n\; (n=1,\ldots,N)$, with $\sum_n s_n=h$.

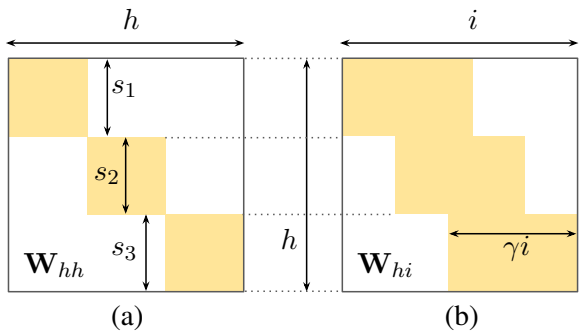
\begin{figure}[!t]
\begin{overpic}[width=\columnwidth]{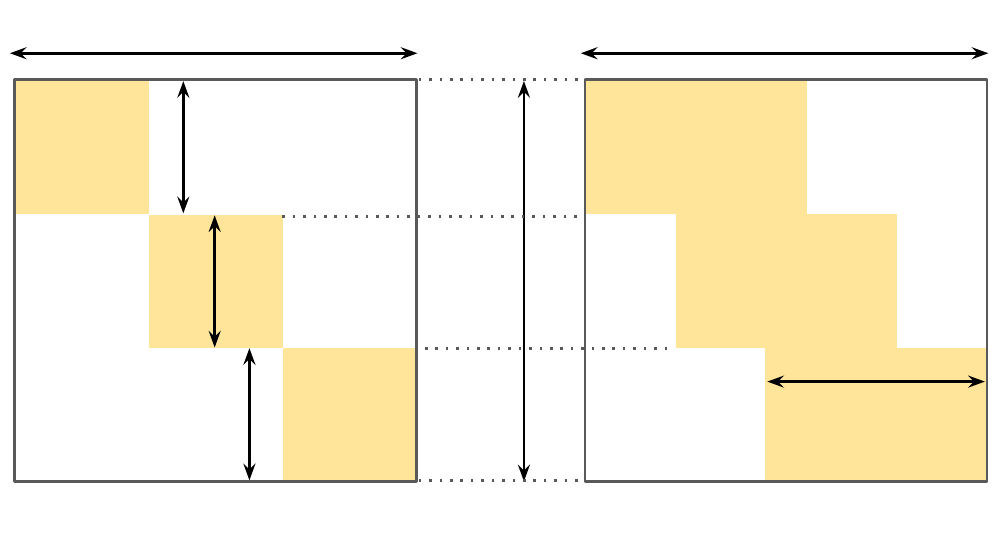}
\put(21,52){$h$}
\put(80,52){$i$}
\put(48,14){$h$}
\put(86,13){$\gamma i$}
\put(4,10){$\mathbf{W}_{hh}$}
\put(61,10){$\mathbf{W}_{hi}$}
\put(19,1){(a)}
\put(76,1){(b)}
\put(19,13){$s_3$}
\put(16,26){$s_2$}
\put(19,41){$s_1$}
\end{overpic}
\caption{Predefined sparseness in hidden-to-hidden ($\mathbf{W}_{hh}$) and input-to-hidden ($\mathbf{W}_{hi}$) matrices in RNNs. %%% recurrent cells.
Trainable parameters (yellow) vs.\ zeros (white).}
\label{fig:rnn_sparse}
\end{figure}

We therefore reduce $\mathbf{W}_{hh}$ to a block-diagonal matrix. For example, a uniform segmentation would reduce the number of trainable parameters in $\mathbf{W}_{hh}$ to a fraction $1/N$. 
\Figref{fig:rnn_sparse} illustrates an example $\mathbf{W}_{hh}$ for $N=3$.
One would expect that this simplification has a significant regularizing effect, given that the number of possible interactions between hidden dimensions is strongly reduced.
However, our experiments (see \secref{subsec:LM}) show that a larger sparse model may still be more expressive than its dense counterpart with the same number of parameters.
Yet, \citet{merity_2017} showed that applying weight dropping (\ie DropConnect, \citet{Wan:2013}) in an LSTM's $\mathbf{W}_{hh}$ matrices  
has a stronger positive effect on language models than other ways to regularize them. Sparsifying $\mathbf{W}_{hh}$ upfront can hence be seen as a similar way to avoid the model's `over-expressiveness' in its recurrent weights.

As a second way to sparsify the RNN cell, we propose to not provide all hidden dimensions with explicit access to each input dimension. In each row of $\mathbf{W}_{hi}$ we limit the number of trainable parameters to a fraction $\gamma\in\ ]0,1]$. 
Practically, we choose to organize the $\gamma i$ trainable parameters in each row within a window that gradually moves from the first to the last input dimension, when advancing in the hidden (i.e., row) dimension.  
Furthermore, we segment the hidden dimension of $\mathbf{W}_{hi}$ according to the segmentation of $\mathbf{W}_{hh}$, and move the window of $\gamma i$ trainable parameters discretely per segment, as illustrated in  \figref{fig:rnn_sparse}(b).

Because of the proposed practical arrangement of sparse and dense blocks in $\mathbf{W}_{hh}$ and $\mathbf{W}_{hi}$, the sparse RNN cell is equivalent to a composition of smaller dense RNN's operating in parallel on (partly) overlapping input data segments, with concatenation of the individual hidden states at the output.  This will be illustrated at the end of \secref{sec:L2R}. 
As a result, fast libraries like CuDNN \cite{Chetlur_2014} can be used directly. Further research is required to investigate the potential benefit in terms of speed and total cell capacity, of physically distributing computations for the individual dense recurrent cells.  

Note that this is only possible because of the initial requirement that the output dimensions are divided into disjoint segments. Whereas inputs can be shared entirely between different components, joining overlapping segments in the $h$ dimension would need to be done within the cell, before applying the gating and output non-linearities. This would make the proposed model less interesting for practical use.

We point out two special cases: 
\begin{enumerate*}[label=(\roman*)]
\item dense $\mathbf{W}_{hi}$ matrices ($\gamma=1$) lead to $N$ parallel RNNs that share the inputs but with separate contributions to the output, and
\item organizing $\mathbf{W}_{hi}$ as a block matrix (e.g., $\gamma=1/N$ for $N$ same-length segments), leads to $N$ isolated parallel RNNs. 
In the latter case, the reduction in trainable parameters is highest, for a given number of segments, but there is no more influence from any input dimension in a given segment to output dimensions in non-corresponding segments. 
\end{enumerate*}
We recommend option (i) as the most rational way to apply our ideas: the sparse RNN output is a concatenation of individual outputs of a number of RNN components connected in parallel, all sharing the entire input.

\subsection{Language Modeling with Sparse RNNs}\label{subsec:LM}
We apply predefined sparse RNNs to language modeling. Our baseline approach is the AWD-LSTM model introduced by \citet{merity_2017}. 
The recurrent unit consists of a three-layer stacked LSTM (Long Short-Term Memory network \cite{hochreiter_97}), with 400-dimensional inputs and outputs, and intermediate hidden state sizes of 1150.
Since the vocabulary contains only $10$k words, most trainable parameters are in the recurrent layer ($20$M out of a total of $24$M). 
In order to cleanly measure the impact of predefined sparseness in the recurrent layer, we maintain the original word embedding layer dimensions, and sparsify the recurrent layer.\footnote{Alternative models could be designed for comparison, with modifications in both the embedding and output layer. Straightforward ideas include an ensemble of smaller independent models, or a mixture-of-softmaxes output layer to combine hidden states of the parallel LSTM components, inspired by \cite{Yang_2017}.}
In this example, we experiment with increasing dimensions in the recurrent layer while maintaining the number of trainable parameters, whereas in \secref{subsec:POS} we increase sparseness while maintaining dimensions. 

Specifically, each LSTM layer is made sparse in such a way that the hidden dimension 1150 is increased by a factor 1.5 (chosen \emph{ad hoc}) to 1725, but the embedding dimensions and total number of parameters remain the same
(within error margins from rounding to integer dimensions for the dense blocks). 
We use uniform segments. 
The number of parameters for the middle LSTM layer can be calculated as:%
\footnote{This follows from an LSTM's 4 $\mathbf{W}_{hh}$ and 4 $\mathbf{W}_{hi}$ matrices, as well as bias vectors. However, depending on the implementation the equations may differ slightly in the contribution from the bias terms. We assume the standard Pytorch implementation~\cite{paszke_2017}.}
\begin{alignat*}{2}
&\text{\# params. LSTM layer 2 } &&\nonumber\\
&\qquad =  4(h_d\ i_d + h_d^2 +2h_d) &&\text{\emph{(dense)}}\nonumber\\
&\qquad = 4N(\frac{h_s}{N}\gamma i_s + \frac{h_s^2}{N^2} + 2\frac{h_s}{N})\quad &&\text{\emph{(sparse)}}\nonumber
\end{alignat*}
in which the first expression represents the general case (e.g., the \emph{dense} case has input and state sizes $i_d = h_d = 1150$),
and the second part is the \emph{sparse} case composed of $N$ parallel LSTMs with input size $\gamma i_s$, and state size $h_s/N$ (with $i_s = h_s = 1725$).
\emph{Dense} and \emph{sparse} variants have the same number of parameters for 
$N=3$ and $\gamma=0.555$. These values are obtained by identifying both expressions. Note that the equality in model parameters for the dense and sparse case holds only approximately due to rounding errors in $(\gamma i_s)$ and $(h_s/N)$. 

\Figref{fig:rnn_sparse} displays $\mathbf{W}_{hh}$ and $\mathbf{W}_{hi}$ for the middle layer, which has close to $11$M parameters out of the total of $24$M in the whole model. 
A dense model with hidden size $h=1725$ would require $46$M parameters, with $24$M in the middle LSTM alone.

\begin{table}[!t]
\centering
%\resizebox{\columnwidth}{!}{%
\begin{tabular}{@{\extracolsep{4pt}}lcc@{}} % trick for spacing between clines
\toprule
& finetune & test perplexity \\ 
\midrule
\cite{merity_2017} & no & $58.8$ \\
baseline & no & $58.8 \pm 0.3$ \\
sparse LSTM & no & $\mathbf{57.9 \pm 0.3}$ \\ 
\midrule
\cite{merity_2017} & yes & $57.3$ \\
baseline & yes & $\mathbf{56.6 \pm 0.2}$ \\
sparse LSTM & yes & $57.0 \pm 0.2$ \\
\bottomrule
\end{tabular}
% }
\caption{Language modeling for PTB \\(mean $\pm$ stdev).  }
%\vspace{-1.7em}
\label{tab:LMresults}
\end{table}

Given the strong hyperparameter dependence of the AWD-LSTM model, and the known issues in objectively evaluating language models \cite{Melis_2017}, we decided to keep all hyperparameters (i.e., dropout rates and optimization scheme) as in the implementation from \citet{merity_2017}\footnote{Our implementation extends \url{https://github.com/salesforce/awd-lstm-lm}.}, including the weight dropping with $p=0.5$ in the sparse $\mathbf{W}_{hh}$ matrices. 
\tabref{tab:LMresults} shows the test perplexity on a processed version~\cite{Mikolov_2010} of the Penn Treebank (PTB) \cite{Marcus_1993}, both with and without the `finetune' step\footnote{The `finetune' step indicates hot-starting the Averaged Stochastic Gradient Descent optimization once more, after convergence in the initial optimization step \cite{merity_2017}.}, displaying mean and standard deviation over 5 %%% runs with different random seeds. 
different runs.
Without finetuning, the sparse model consistently performs around 1 perplexity point better, whereas after finetuning, the original remains slightly better, although less consistently so over different random seeds. We observed that the sparse model overfits more strongly than the baseline, especially during the finetune step. 
We hypothesize that the regularization effect of \textit{a priori} limiting interactions between dimensions does not compensate for the increased expressiveness of the model due to the larger hidden state size. 
Further experimentation, with tuned hyperparameters, is needed to determine the actual benefits of predefined sparseness, in terms of model size, resulting perplexity, and sensitivity to the choice of hyperparameters.

%\section{Results and Discussion}

\section{Sparse Word Embeddings}\label{sec:sparse_emb}

Given a vocabulary with $V$ words, we want to construct vector representations of length $k$ for each word such that the total number of parameters needed (i.e., non-zero entries), is smaller than $k\, V$. 
We introduce one way to do this based on word frequencies (\secref{subsec:freqbased_sparse_emb}), and present part-of-speech tagging experiments (\secref{subsec:POS}).

\subsection{Word-Frequency based Embedding Size}\label{subsec:freqbased_sparse_emb}

Predefined sparseness in word embeddings amounts to deciding which positions in the word embedding matrix $\mathbf{E}\in \mathbb{R}^{V \times k}$ should be fixed to zero, prior to training.
We define the fraction of trainable entries in $\mathbf{E}$ as the embedding density $\delta_E$.
We hypothesize that rare words can be represented with fewer parameters than frequent words, since they only appear in very specific contexts.  This will be investigated experimentally in \secref{subsec:POS}. 
Word occurrence frequencies have a typical Zipfian nature \cite{Manning_2008}, with many rare and few highly frequent terms.
Thus, representing the long tail of rare terms with short embeddings should greatly reduce memory requirements.

In the case of a low desired embedding density $\delta_E$, we want to save on the rare words, in terms of assigning trainable parameters, and focus on the fewer more popular words. An exponential decay in the number of words that are assigned longer representations is one possible way to implement this. In other words, we propose to have the number of words that receive a trainable parameter at dimension $j$ decrease with a factor $\alpha^j$ ($\alpha \in\ ]0, 1]$).
For a given fraction $\delta_E$, the parameter $\alpha$ can be determined from requiring the total number of non-zero embedding parameters to amount to a given fraction $\delta_E$ of all parameters:
\begin{align}
\text{\# embedding params.} 
= \sum_{j=0}^{k-1}\alpha^j V =  \delta_E\, k\, V \nonumber
\end{align}
and numerically solving for $\alpha$.

\begin{figure}%[!ht]
\includegraphics[width=.9\columnwidth]{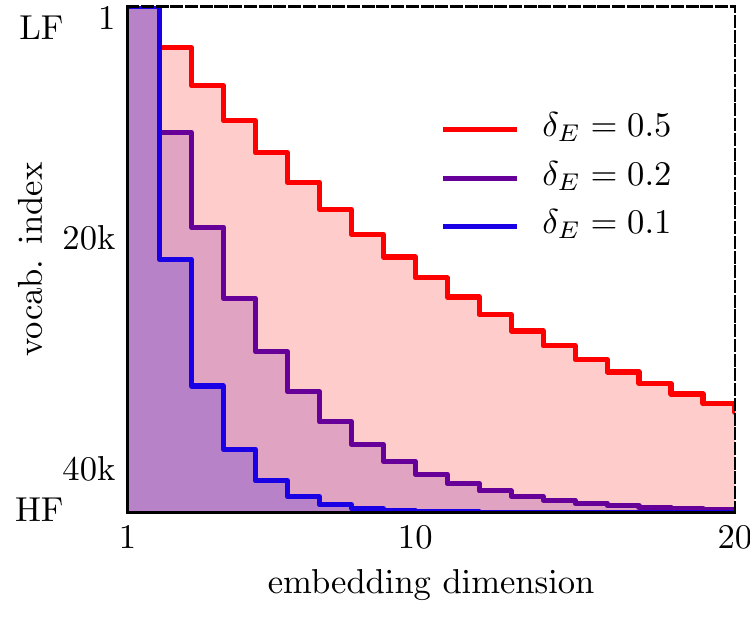}
\caption{Visualization of sparse embedding matrices for different densities $\delta_E$ (with $k = 20$). Colored region: non-zero entries. 
Rows represent word indices, sorted from least frequent (LF) to highly frequent (HF).
}
\label{fig:embmatrix}
\end{figure}

\Figref{fig:embmatrix} gives examples of embedding matrices with varying~$\delta_E$.
For a vocabulary of $44$k terms and maximum embedding length $k=20$, the density $\delta_E=0.2$ leads to 25\% of the words with embedding length 1 (corresponding $\alpha=0.75$), only 7.6\% with length of 10 or higher, and with the maximum length 20 for only the 192 most frequent terms.  
The particular configurations shown in \figref{fig:embmatrix} are used for the experiments in \secref{subsec:POS}.

In order to set a minimum embedding length for the rarest words, as well as for computational efficiency, we note that this strategy can also be applied on $M$ bins of embedding dimensions, rather than per individual dimensions. The width of the first bin then indicates the minimum embedding length. Say bin $m$ has width $\kappa_m$ (for $m=0,\ldots,M-1$, and $\sum_m \kappa_m = k$). The multiplicative decay factor $\alpha$ can then be obtained by solving
\begin{equation}
\delta_E = \frac{1}{k}\sum_{m=0}^{M-1}\kappa_m\alpha^m, \label{eq:deltaE}
\end{equation}
while numerically compensating for rounding errors in the number $V\alpha^m$ of words that are assigned trainable parameters in the $m^\textrm{th}$ bin.

$ $

\subsection{Part-of-Speech Tagging Experiments}\label{subsec:POS}
\begin{figure}%[!ht]
\includegraphics[width=.9\columnwidth]{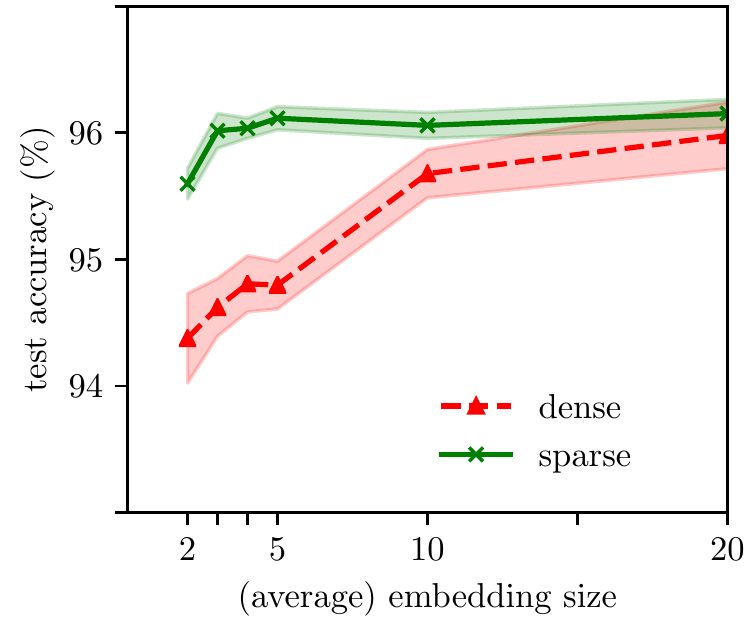}
\caption{POS tagging accuracy on PTB data: dense (red) vs.\ sparse (green). X-axis: embedding size $k$ for the dense case, and average embedding size (or $20\;\delta_E$) for the sparse case. 
Shaded bands indicate \emph{stdev} over 4 randomly seeded runs.
}
\label{fig:pos_sparse}
\end{figure}

We now study the impact of sparseness in word embeddings, for a basic POS tagging model, and report results on the PTB Wall Street Journal data. 
We embed 43,815 terms in 20-dimensional space, as input for a BiLSTM layer with hidden state size 10 
for both forward and backward directions.
The concatenated hidden states go into a fully connected layer with $\tanh$ non-linearity (down to dimension 10), followed by a \emph{softmax} classification layer with 49 outputs (i.e., the number of POS tags). 
The total number of parameters is $880$k, of which $876$k in the embedding layer.
Although character-based models are known to outperform pure word embedding based models \cite{Ling_2015}, we wanted to investigate the effect of sparseness in word embeddings, rather than creating more competitive but larger or complex models, risking a smaller resolution in the effect of changing
individual building blocks. To this end we also limited the dimensions, and hence the expressiveness, of the recurrent layer.\footnote{With LSTM state sizes of 50, the careful tuning of dropout parameters gave an accuracy of 94.7\% when reducing the embedding size to $k=2$, a small gap compared to 96.8\% for embedding size 50. The effect of larger sparse embeddings was therefore much smaller in absolute value than the one visualized in \figref{fig:pos_sparse}, because of the much more expressive recurrent layer.} 
Our model is similar to but smaller than the `word lookup' baseline by \citet{Ling_2015}. 
%, although with reduced embedding and RNN sizes 

\Figref{fig:pos_sparse} compares the accuracy for variable densities $\delta_E$ (for $k=20$) vs.\ different embedding sizes (with $\delta_E=1$).
For easily comparing sparse and dense models with the same number of embedding parameters, we scale $\delta_E$, the x-axis for the sparse case, to the average embedding size of $20\;\delta_E$.

Training models with shorter dense embeddings appeared more difficult. In order to make a fair comparison, we therefore tuned the models over a range of regularization hyperparameters, provided in \tabref{tab:POSparams}.

\begin{table*}[t]
\centering
%\resizebox{\columnwidth}{!}{%
\begin{tabular}{ll} % trick for spacing between clines
\toprule
hyperparameter & value(s) \\
\midrule
optimizer & Adam \cite{kingma_2014} \\
learning rate & 0.001 \\
epochs & 50 \\
word level embedding dropout \dag & [0.0, 0.1, 0.2] \\
variational embedding dropout \dag & [0.0, 0.1, 0.2, 0.4] \\
DropConnect on $\mathbf{W}_{hh}$ \dag & [0.0, 0.2, 0.4]\\
batch size & 20 \\
\bottomrule
\end{tabular}
% }
 \caption{Hyperparameters for POS tagging model (\dag as introduced in \cite{merity_2017}). A list indicates tuning over the given values was performed.}
%\vspace{-1.7em}
\label{tab:POSparams}
 \end{table*}

We observe that the sparse embedding layer allows lowering the number of parameters in $\mathbf{E}$ down to a fraction of 15\% of the original amount, with little impact on the effectiveness, provided $\mathbf{E}$ is sparsified rather than reduced in size. 
The reason for that is that with sparse 20-dimensional embeddings, the BiLSTM still receives 20-dimensional inputs, from which a significant subset only transmits signals from a small set of frequent terms. In the case of smaller dense embeddings, information from all terms is uniformly present over fewer dimensions, and needs to be processed with fewer parameters at the encoder input.

Finally, we verify the validity of our hypothesis from \secref{subsec:freqbased_sparse_emb} that frequent terms need to be embedded with more parameters than rare words.
Indeed, one could argue in favor of the opposite strategy. It would be computationally more efficient if the terms most often encountered had the smallest representation. Also, stop words are the most frequent ones but are said to carry little information content. 
However, \tabref{tab:POSresults} confirms our initial hypothesis. Applying the introduced strategy to sparsify embeddings on randomly ordered words (`no sorting') leads to a significant decrease in accuracy compared to the proposed sorting strategy (`up').
When the most frequent words are encoded with the shortest embeddings (`down' in the table), the accuracy goes down even further. 
\begin{table*}[t!]
\centering
%\resizebox{\columnwidth}{!}{%
\begin{tabular}{@{\extracolsep{4pt}}lccc@{}} % trick for spacing between clines
\toprule
& $\delta_E=1.0$ & $\delta_E=0.25$ & $\delta_E=0.1$ \\ 
\midrule
\# params. ($\mathbf{E}$; all) & $876$k; $880$k  & $219$k; $222$k  & $88$k ; $91$k  \\
%\# params (total)         & $880$k & $222$k & $91$k \\
%\# params ($\mathbf{E}$)  & $876$k & $219$k& $88$k \\
\midrule
up          &                & $\mathbf{96.1 \pm 0.1}$ & $\mathbf{95.6 \pm 0.1}$ \\
no sorting  & $96.0 \pm 0.3$ & $94.3 \pm 0.4$ & $93.0 \pm 0.3$ \\
down        &                & $89.8 \pm 2.2$ & $90.6 \pm 0.5$ \\
\bottomrule
\end{tabular}
% }
 \caption{Impact of vocabulary sorting on POS accuracy with sparse embeddings: up vs.\ down (most frequent words get longest vs.\ shortest embeddings, resp.) or not sorted, for different embedding densities $\delta_E$.
}
%\vspace{-1.7em}
\label{tab:POSresults}
 \end{table*}

\section{Learning To Recite}\label{sec:L2R}

\begin{table*}[t!]
\centering
\begin{tabular}{@{\extracolsep{4pt}}lccccc@{}} % trick for spacing between clines
\toprule
&\multicolumn{2}{c}{embeddings} & hidden state & \multirow{2}{*}{\# parameters} & memorization \\
& size $k$, & density $\delta_E$ & size $h$ &  & accuracy (\%) \\
\midrule
dense model  (orig. dims.)  & $400$ &  $1$  &  $1150$  &  $24.22$M  & $\mathbf{100.0}$  \\
\midrule
dense model (see \figref{fig:rnn_composite}(a))                & $200$ & $1$ & $575$     & $7.07$M  &  $99.33$ \\
sparse RNN  (see \figref{fig:rnn_composite}(b))                 & $200$ & $1$ & $1150$     & $7.07$M  & $\mathbf{99.95}$  \\
sparse RNN + sparse emb.    & $400$ & $1/2$ & $1150$     & $7.07$M  & $99.74$  \\
\midrule
dense model                 & $133$ & $1$ & $383$     & $3.59$M  & $81.48$  \\
sparse RNN                  & $133$ & $1$ & $1150$    & $3.59$M  & $76.37$  \\
sparse RNN + sparse emb.    & $400$ & $1/3$ & $1150$     & $3.59$M  & $\mathbf{89.98}$  \\
\bottomrule
\end{tabular}
 \caption{PTB train set memorization accuracies for dense models vs.\ models with predefined sparseness in recurrent and embedding layers with comparable number of parameters.}
%\vspace{-1.7em}
\label{tab:L2R}
 \end{table*}

From the language modeling experiments in \secref{subsec:LM}, we hypothesized that an RNN layer becomes more expressive, when the dense layer is replaced by a larger layer with predefined sparseness and the same number of model parameters. In this section, we design an experiment to further investigate this claim. One way of quantifying an RNN's capacity is in measuring how much information it can memorize. We name our experimental setup \emph{learning to recite}: we investigate to what extent dense vs.\ sparse models are able to learn an entire corpus by heart in order to recite it afterwards.
We note that this toy problem could have interesting applications, such as the design of neural network components that keep entire texts or even knowledge bases available for later retrieval, encoded in the component's weight matrices.\footnote{It is likely that recurrent networks are not the best choice for this purpose, but here we only wanted to measure the LSTM-based model's capacity to memorize with and without predefined sparseness.}
%\todo{cite memorizing hash table with NN?}

\subsection{Experimental Results}
The initial model for our \emph{learning to recite} experiment is the baseline language model used in \secref{subsec:LM} \cite{merity_2017}, with the PTB data. We set all regularization parameters to zero, to focus on memorizing the training data. During training, we measure the ability of the model to correctly predict the next token at every position in the training data, by selecting the token with highest predicted probability. When the model reaches an accuracy of 100\%, it is able to recite the entire training corpus. 
We propose the following optimization setup (tuned and tested on dense models with different sizes):
minibatch SGD (batch size 20, momentum 0.9, and best initial learning rate among 5 or 10). An exponentially decaying learning rate factor (0.97 every epoch) appeared more suitable for memorization than other learning rate scheduling strategies, and we report the highest accuracy in 150 epochs.

We compare the original model (in terms of network dimensions) with versions that have less parameters,
by either reducing the RNN hidden state size $h$ or by sparsifying the RNN, and similarly for the embedding layer. For making the embedding matrix sparse, $M=10$ equal-sized segments are used (as in eq.~\ref{eq:deltaE}).
\Tabref{tab:L2R} lists the results. 
The dense model with the original dimensions has $24$M parameters to memorize a sequence of in total `only' $930$k tokens, and is able to do so. When the model's embedding size and intermediate hidden state size are halved, the number of parameters drops to $7$M, and the resulting model now makes 67 mistakes out of 10k predictions. If $h$ is kept, but the recurrent layers are made sparse to yield the same number of parameters, only 5 mistakes are made for every 10k predictions. Making the embedding layer sparse as well introduces new errors. 
If the dimensions are further reduced to a third the original size, the memorization capacity goes down strongly, with less than $4$M trainable parameters. In this case, sparsifying both the recurrent and embedding layer yields the best result, whereas the dense model works better than the model with sparse RNNs only. 
A possible explanation for that is the strong sparseness in the RNNs. For example, in the middle layer only 1 out of 10 recurrent connections is non-zero. In this case, increasing the size of the word embeddings (at least, for the frequent terms) could provide an alternative for the model to memorize parts of the data, or maybe it makes the optimization process more robust.

\begin{figure}[!t]
\begin{overpic}[width=\columnwidth]{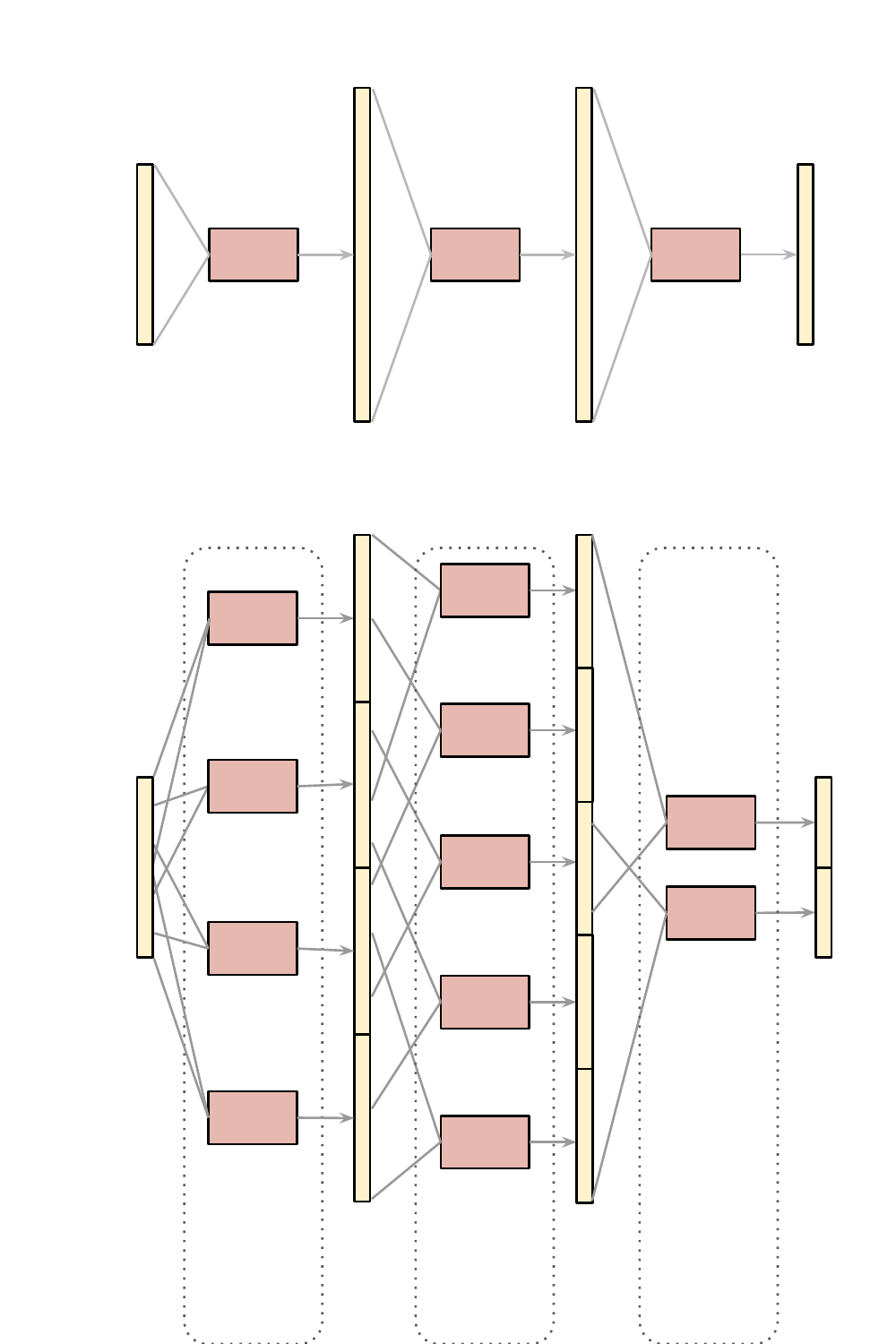}
%\begin{overpic}[width=\columnwidth, grid,ticks=5]{sparse_rnn_composed.pdf}
%
\put(16,98){\small{layer 1}}
\put(32,98){\small{layer 2}}
\put(48,98){\small{layer 3}}
%figure (a)
\put(0,82){(a)}
\put(4,89){\small{$k\!=\!200$}}
\put(22,95){\small{$h\!=\!575$}}
\put(40,95){\small{$h\!=\!575$}}
\put(57,89){\small{$k\!=\!200$}}
\put(16,80){\footnotesize{$R_1$}}
\put(32,80){\footnotesize{$R_2$}}
\put(49,80){\footnotesize{$R_3$}}
\put(14,76){\scriptsize{$200\!\rightarrow\!575$}}
\put(31,76){\tiny{$575\!\rightarrow\!575$}}
\put(48,76){\tiny{$575\!\rightarrow\!200$}}
\put(14,73){\tiny{$1.79$M par.}}
\put(31,73){\tiny{$2.65$M par.}}
\put(48,73){\tiny{$0.62$M par.}}
%
%figure (b)
\put(0,35){(b)}
\put(4,44){\small{$k\!=\!200$}}
\put(22,62){\small{$h\!=\!1150$}}
\put(40,62){\small{$h\!=\!1150$}}
\put(57,44){\small{$k\!=\!200$}}
\put(16,53){\footnotesize{$R_{1,1}$}}
\put(16,41){\footnotesize{$R_{1,2}$}}
\put(16,29){\footnotesize{$R_{1,3}$}}
\put(16,16){\footnotesize{$R_{1,4}$}}
\put(33,55){\footnotesize{$R_{2,1}$}}
\put(33,45){\footnotesize{$R_{2,2}$}}
\put(33,35){\footnotesize{$R_{2,3}$}}
\put(33,24.5){\footnotesize{$R_{2,4}$}}
\put(33,14){\footnotesize{$R_{2,5}$}}
\put(50,38){\footnotesize{$R_{3,1}$}}
\put(50,31){\footnotesize{$R_{3,2}$}}
\put(14,12){\tiny{$99\!\rightarrow\!288$}}
\put(31,10){\tiny{$244\!\rightarrow\!230$}}
\put(48,27){\tiny{$675\!\rightarrow\!100$}}
\put(14,6){\footnotesize{sparse $R_{1}$}}
\put(14,3){\tiny{$200\!\rightarrow\!1150$}}
\put(31,6){\footnotesize{sparse $R_{2}$}}
\put(31,3){\tiny{$1150\!\rightarrow\!1150$}}
\put(48,6){\footnotesize{sparse $R_{3}$}}
\put(48,3){\tiny{$1150\!\rightarrow\!200$}}

\put(14,0){\tiny{$1.79$M par.}}
\put(31,0){\tiny{$2.65$M par.}}
\put(48,0){\tiny{$0.62$M par.}}

%\put(55,10){\includegraphics[scale=.07]%
%{golfer.ps}}
\end{overpic}
\caption{Schematic overview of 3-layer stacked (a) dense vs.\ (b) sparse LSTMs with the same number of parameters (indicated with `par.'). Sparse layers are effectively composed of smaller dense LSTMs. `$R_{i,j}$' indicates component $j$ within layer $i$, and `$675\!\rightarrow\!100$' indicates an LSTM compoment with input size $675$ and output size $100$.}
\label{fig:rnn_composite}
\end{figure}
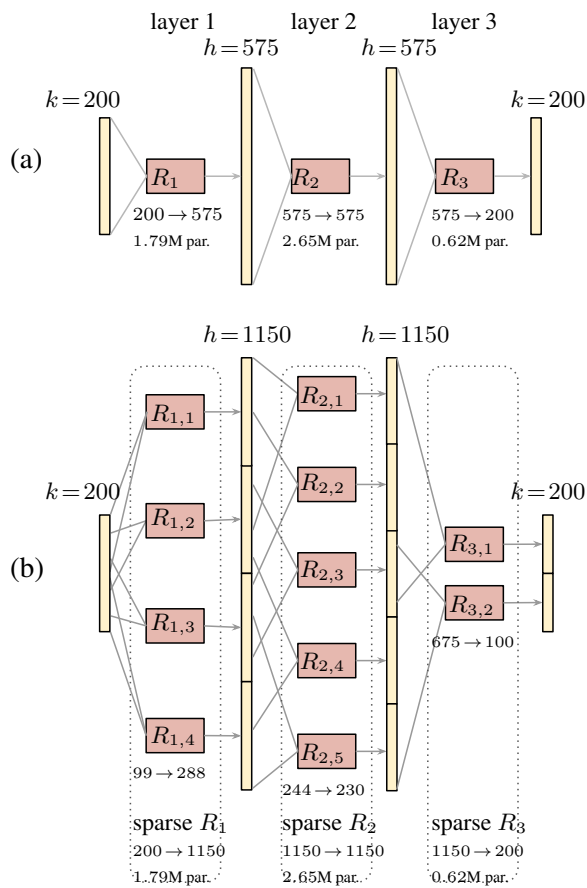

\subsection{Visualization}
Finally, we provide an illustration of the high-level composition of the recurrent layers in two of the models used for this  experiment. \Figref{fig:rnn_composite}(a) sketches the stacked 3-layer LSTM network from the `dense RNN' model (see \Tabref{tab:L2R}) with $k=200$ and $h=575$.
As already mentioned, our proposed sparse LSTMs are equivalent to a well-chosen composition of smaller dense LSTM components with overlapping inputs and disjoint outputs. This composition is shown in \figref{fig:rnn_composite}(b) for the model `sparse RNN'  (see \Tabref{tab:L2R}), which in every layer has the same number of parameters as the dense model with reduced dimensions.

\section{Conclusion and Future Work}\label{sec:conclusion}
This paper introduces strategies to design word embedding layers and recurrent networks with predefined sparseness. Effective sparse word representations can be constructed by 
encoding less frequent terms with smaller embeddings and vice versa.  A sparse recurrent neural network layer can be constructed by applying multiple smaller recurrent cells in parallel, with partly overlapping inputs and concatenated outputs. 

The presented ideas can be applied to build models with larger representation sizes 
for a given number of parameters, as illustrated with a language modeling example.
Alternatively, they can be used 
to reduce the number of parameters %compared to the dense case, 
for given representation sizes, as investigated with a part-of-speech tagging model.

We introduced ideas on predefined sparseness in sequence models, as well as proof-of-concept experiments, and analysed the memorization capacity of sparse networks in the `learning to recite' toy problem.

More elaborate experimentation is required to investigate the benefits of predefined sparseness on more competitive tasks and datasets in NLP. For example, language modeling results on the Penn Treebank rely on heavy regularization due to the small corpus. Follow-up work could therefore investigate to what extent language models for large corpora can be trained with limited computational resources, based on predefined sparseness. 
Other ideas for future work include the use
of predefined sparseness for pretraining word embeddings,
or other neural network components besides
recurrent models, as well as their use in alternative
applications such as sequence-to-sequence
tasks or in multi-task scenarios.

\section*{Acknowledgments}
We thank the anonymous reviewers for their time and effort, and the valuable feedback.

%\bibliography{conll2018}

\begin{thebibliography}{32}
\expandafter\ifx\csname natexlab\endcsname\relax\def\natexlab#1{#1}\fi

\bibitem[{Bekoulis et~al.(2018)Bekoulis, Deleu, Demeester, and
  Develder}]{bekoulis2018}
Giannis Bekoulis, Johannes Deleu, Thomas Demeester, and Chris Develder. 2018.
\newblock Joint entity recognition and relation extraction as a multi-head
  selection problem.
\newblock \emph{Expert Systems with Applications}, 114:34--45.

\bibitem[{Chelba et~al.(2013)Chelba, Mikolov, Schuster, Ge, Brants, Koehn, and
  Robinson}]{Chelba2013}
Ciprian Chelba, Tomas Mikolov, Mike Schuster, Qi~Ge, Thorsten Brants, Phillipp
  Koehn, and Tony Robinson. 2013.
\newblock One billion word benchmark for measuring progress in statistical
  language modeling.
\newblock Technical report, Google.

\bibitem[{Chetlur et~al.(2014)Chetlur, Woolley, Vandermersch, Cohen, Tran,
  Catanzaro, and Shelhamer}]{Chetlur_2014}
Sharan Chetlur, Cliff Woolley, Philippe Vandermersch, Jonathan Cohen, John
  Tran, Bryan Catanzaro, and Evan Shelhamer. 2014.
\newblock {cuDNN}: Efficient primitives for deep learning.
\newblock \emph{arXiv:1410.0759}.

\bibitem[{Chung et~al.(2014)Chung, G{\"{u}}l{\c c}ehre, Cho, and
  Bengio}]{Chung2014_GRU}
Junyoung Chung, {\c C}ağlar G{\"{u}}l{\c c}ehre, Kyunghyun Cho, and Yoshua
  Bengio. 2014.
\newblock Empirical evaluation of gated recurrent neural networks on sequence
  modeling.
\newblock \emph{arXiv:1412.3555}.
\newblock Deep Learning workshop at NIPS 2014.

\bibitem[{Goodfellow et~al.(2016)Goodfellow, Bengio, and
  Courville}]{Goodfellow_2016}
Ian Goodfellow, Yoshua Bengio, and Aaron Courville. 2016.
\newblock \emph{Deep Learning}.
\newblock MIT Press.
\newblock \url{http://www.deeplearningbook.org}.

\bibitem[{Guo et~al.(2016)Guo, Yao, and Chen}]{guo_dynamic_2016}
Yiwen Guo, Anbang Yao, and Yurong Chen. 2016.
\newblock Dynamic network surgery for efficient {DNNs}.
\newblock In \emph{Proc. 30th {International} {Conference} on {Neural}
  {Information} {Processing} {Systems} (NIPS 2016)}, {NIPS}'16, pages
  1387--1395.

\bibitem[{Han et~al.(2016)Han, Mao, and Dally}]{han_deep_2015}
Song Han, Huizi Mao, and William~J. Dally. 2016.
\newblock Deep compression: Compressing deep neural networks with pruning,
  trained quantization and {H}uffman coding.
\newblock In \emph{Proc. 4th International Conference on Learning
  Representations (ICLR 2016)}.

\bibitem[{Han et~al.(2015)Han, Pool, Tran, and Dally}]{han_learning_2015}
Song Han, Jeff Pool, John Tran, and William~J. Dally. 2015.
\newblock Learning {Both} {Weights} and {Connections} for {Efficient} {Neural}
  {Networks}.
\newblock In \emph{Proc. 28th International Conference on Neural Information
  Processing Systems (NIPS 2015)}, {NIPS}'15, pages 1135--1143.

\bibitem[{Hashimoto et~al.(2017)Hashimoto, Xiong, Tsuruoka, and
  Socher}]{hashimoto2017}
Kazuma Hashimoto, Caiming Xiong, Yoshimasa Tsuruoka, and Richard Socher. 2017.
\newblock A joint many-task model: Growing a neural network for multiple nlp
  tasks.
\newblock In \emph{Proc. Conference on Empirical Methods in Natural Language
  Processing (EMNLP)}, pages 1923--1933.

\bibitem[{Hochreiter and Schmidhuber(1997)}]{hochreiter_97}
Sepp Hochreiter and J\"{u}rgen Schmidhuber. 1997.
\newblock Long short-term memory.
\newblock \emph{Neural computation}, 9(8):1735--1780.

\bibitem[{Jaderberg et~al.(2014)Jaderberg, Vedaldi, and
  Zisserman}]{jaderberg_speeding_2014}
Max Jaderberg, Andrea Vedaldi, and Andrew Zisserman. 2014.
\newblock Speeding up convolutional neural networks with low rank expansions.
\newblock In \emph{Proc. 27th British Machine Vision Conference (BMVC 2014)}.
\newblock ArXiv: 1405.3866.

\bibitem[{Kalchbrenner et~al.(2018)Kalchbrenner, Elsen, Simonyan, Noury,
  Casagrande, Lockhart, Stimberg, van~den Oord, Dieleman, and
  Kavukcuoglu}]{kalchbrenner_2018}
Nal Kalchbrenner, Erich Elsen, Karen Simonyan, Seb Noury, Norman Casagrande,
  Edward Lockhart, Florian Stimberg, A{\"{a}}ron van~den Oord, Sander Dieleman,
  and Koray Kavukcuoglu. 2018.
\newblock Efficient neural audio synthesis.
\newblock ArXiv: 1802.08435.

\bibitem[{Kingma and Ba(2015)}]{kingma_2014}
Diederik Kingma and Jimmy Ba. 2015.
\newblock {A}dam: {A} method for stochastic optimization.
\newblock In \emph{International Conference on Learning Representations}, San
  Diego, USA.

\bibitem[{Lebedev and Lempitsky(2016)}]{lebedev_fast_2016}
V.~Lebedev and V.~Lempitsky. 2016.
\newblock Fast {ConvNets} using group-wise brain damage.
\newblock In \emph{Proc. 29th IEEE Conference on Computer Vision and Pattern
  Recognition (CVPR 2016)}, pages 2554--2564.

\bibitem[{Ling et~al.(2015)Ling, Dyer, Black, Trancoso, Fermandez, Amir,
  Marujo, and Luis}]{Ling_2015}
Wang Ling, Chris Dyer, Alan~W Black, Isabel Trancoso, Ramon Fermandez, Silvio
  Amir, Luis Marujo, and Tiago Luis. 2015.
\newblock Finding function in form: Compositional character models for open
  vocabulary word representation.
\newblock In \emph{Proceedings of the 2015 Conference on Empirical Methods in
  Natural Language Processing}, pages 1520--1530, Lisbon, Portugal. Association
  for Computational Linguistics.

\bibitem[{Liu et~al.(2015)Liu, Wang, Foroosh, Tappen, and
  Penksy}]{liu_sparse_2015}
Baoyuan Liu, Min Wang, H.~Foroosh, M.~Tappen, and M.~Penksy. 2015.
\newblock Sparse convolutional neural networks.
\newblock In \emph{Proc. 28th IEEE Conference on Computer Vision and Pattern
  Recognition (CVPR 2015)}, pages 806--814.

\bibitem[{Lu et~al.(2016)Lu, Sindhwani, and Sainath}]{lu_learning_2016}
Zhiyun Lu, Vikas Sindhwani, and Tara~N. Sainath. 2016.
\newblock Learning compact recurrent neural networks.
\newblock In \emph{Proc. 41st IEEE International Conference on Acoustics,
  Speech and Signal Processing (ICASSP 2016)}.

\bibitem[{Manning et~al.(2008)Manning, Raghavan, and
  Sch\"{u}tze}]{Manning_2008}
Christopher~D. Manning, Prabhakar Raghavan, and Hinrich Sch\"{u}tze. 2008.
\newblock \emph{Introduction to Information Retrieval}.
\newblock Cambridge University Press, New York, NY, USA.

\bibitem[{Marcus et~al.(1993)Marcus, Santorini, and
  Marcinkiewicz}]{Marcus_1993}
Mitchell~P. Marcus, Beatrice Santorini, and Mary~Ann Marcinkiewicz. 1993.
\newblock Building a large annotated corpus of english: The penn treebank.
\newblock \emph{Computational Linguistics}, 19(2):313--330.

\bibitem[{Melis et~al.(2017)Melis, Dyer, and Blunsom}]{Melis_2017}
Gábor Melis, Chris Dyer, and Phil Blunsom. 2017.
\newblock On the state of the art of evaluation in neural language models.
\newblock In \emph{Proc. 6th International Conference on Learning
  Representations (ICLR 2017)}.

\bibitem[{Merity et~al.(2017)Merity, Keskar, and Socher}]{merity_2017}
Stephen Merity, Nitish~Shirish Keskar, and Richard Socher. 2017.
\newblock Regularizing and optimizing {LSTM} language models.
\newblock \emph{arXiv:1708.02182}.

\bibitem[{Mikolov et~al.(2010)Mikolov, Karafiát, Burget, Cernocký, and
  Khudanpur}]{Mikolov_2010}
Tomas Mikolov, Martin Karafiát, Lukás Burget, Jan Cernocký, and Sanjeev
  Khudanpur. 2010.
\newblock Recurrent neural network based language model.
\newblock In \emph{INTERSPEECH}, pages 1045--1048. ISCA.

\bibitem[{Miwa and Bansal(2016)}]{Miwa2016}
Makoto Miwa and Mohit Bansal. 2016.
\newblock End-to-end relation extraction using {LSTMs} on sequences and tree
  structures.
\newblock In \emph{Proc. 54th Annual Meeting of the Association for
  Computational Linguistics}, pages 1105--1116.

\bibitem[{Molchanov et~al.(2017)Molchanov, Ashukha, and
  Vetrov}]{molchanov_variational_2017}
Dmitry Molchanov, Arsenii Ashukha, and Dmitry Vetrov. 2017.
\newblock Variational dropout sparsifies deep neural networks.
\newblock In \emph{Proc. 35th International Conference on Machine Learning
  (ICML 2017)}.
\newblock ArXiv: 1701.05369.

\bibitem[{Narang et~al.(2017)Narang, Elsen, Diamos, and
  Sengupta}]{narang_exploring_2017}
Sharan Narang, Erich Elsen, Gregory Diamos, and Shubho Sengupta. 2017.
\newblock Exploring sparsity in recurrent neural networks.
\newblock In \emph{Proc. 5th International Conference on Learning
  Representations (ICLR 2017)}.

\bibitem[{Paszke et~al.(2017)Paszke, Gross, Chintala, Chanan, Yang, DeVito,
  Lin, Desmaison, Antiga, and Lerer}]{paszke_2017}
Adam Paszke, Sam Gross, Soumith Chintala, Gregory Chanan, Edward Yang, Zachary
  DeVito, Zeming Lin, Alban Desmaison, Luca Antiga, and Adam Lerer. 2017.
\newblock Automatic differentiation in pytorch.
\newblock In \emph{Proceedings of the Workshop on The future of gradient-based
  machine learning software and techniques, co-located with the 31st Annual
  Conference on Neural Information Processing Systems (NIPS 2017)}.

\bibitem[{Shazeer et~al.(2017)Shazeer, Mirhoseini, Maziarz, Davis, Le, Hinton,
  and Dean}]{Shazeer2017}
Noam Shazeer, Azalia Mirhoseini, Krzysztof Maziarz, Andy Davis, Quoc Le,
  Geoffrey Hinton, and Jeff Dean. 2017.
\newblock Outrageously large neural networks: The sparsely-gated
  mixture-of-experts layer.
\newblock In \emph{Proc. International Conference on Learning Representations
  (ICLR)}.

\bibitem[{Tai et~al.(2016)Tai, Xiao, Zhang, Wang, and
  E}]{tai_convolutional_2015}
Cheng Tai, Tong Xiao, Yi~Zhang, Xiaogang Wang, and Weinan E. 2016.
\newblock Convolutional neural networks with low-rank regularization.
\newblock In \emph{Proc. 4th International Conference on Learning
  Representations (ICLR 2016)}.
\newblock ArXiv: 1511.06067.

\bibitem[{Wan et~al.(2013)Wan, Zeiler, Zhang, LeCun, and Fergus}]{Wan:2013}
Li~Wan, Matthew Zeiler, Sixin Zhang, Yann LeCun, and Rob Fergus. 2013.
\newblock Regularization of neural networks using dropconnect.
\newblock In \emph{Proc. 30th International Conference on International
  Conference on Machine Learning (ICML 2013)}, pages III--1058--III--1066,
  Atlanta, GA, USA.

\bibitem[{Wang et~al.(2016)Wang, Zhou, and Hussain}]{wang_deep_2016}
Dong Wang, Qiang Zhou, and Amir Hussain. 2016.
\newblock Deep and sparse learning in speech and language processing: An
  overview.
\newblock In \emph{Proc. 8th International Conference on (BICS2016)}, pages
  171--183. Springer, Cham.

\bibitem[{Wen et~al.(2016)Wen, Wu, Wang, Chen, and Li}]{wen_learning_2016}
Wei Wen, Chunpeng Wu, Yandan Wang, Yiran Chen, and Hai Li. 2016.
\newblock Learning structured sparsity in deep neural networks.
\newblock In \emph{Proc. 30th International Conference on Neural Information
  Processing Systems (NIPS 2016)}, {NIPS}'16, pages 2082--2090, USA.

\bibitem[{Yang et~al.(2017)Yang, Dai, Salakhutdinov, and Cohen}]{Yang_2017}
Zhilin Yang, Zihang Dai, Ruslan Salakhutdinov, and William~W. Cohen. 2017.
\newblock Breaking the softmax bottleneck: A high-rank rnn language model.
\newblock ArXiv: 1711.03953.

\end{thebibliography}
%\bibliographystyle{acl_natbib_nourl}

\clearpage 

%\appendix

%\section{Supplemental Material}
%\label{sec:supplemental}

\end{document}